\documentclass{article} 
\usepackage{tencent_tech_report}
\usepackage[colorlinks = true,
            linkcolor = blue,
            urlcolor  = blue,
            citecolor = blue,
            anchorcolor = blue]{hyperref}   

\usepackage{tcolorbox} 
\newtcolorbox{alprompt}[1]{
        boxrule = 1pt,
        fontupper = \small\tt,
        fonttitle = \bf\color{black},
        arc = 2pt,
        rounded corners,
        colframe = black,
        colbacktitle = white!97!yellow,
        colback = white!97!yellow,
        title = #1,
}
\newtcolorbox{promptbox}[3][Prompt]{
colback=black!5!white,
arc=5pt, 
boxrule=0.5pt,
fonttitle=\bfseries,
title=#1, 
before upper={\small}, fontupper=\fontfamily{ptm}\selectfont,
colframe=#2,
label=#3,
}

\usepackage{microtype}
\usepackage{hyperref}
\usepackage{url}
\usepackage{booktabs}
\usepackage{enumitem}
\usepackage{multicol}
\usepackage{multirow}
\usepackage{makecell}
\usepackage{CJKutf8}
\usepackage{amsmath}
\usepackage{siunitx}
\usepackage{floatflt}
\usepackage{wrapfig}
\usepackage{graphicx}
\usepackage{booktabs}
\usepackage{wrapfig}
\usepackage{authblk}
\usepackage{lipsum}

\usepackage{algorithm}
\usepackage{algorithmicx}
\usepackage{algpseudocode}
\usepackage{microtype}
\usepackage{graphicx}
\usepackage{multirow}
\usepackage{booktabs} 
\usepackage{pifont}  
\usepackage{graphicx}  
\usepackage{subcaption} 
\usepackage{hyperref}

\usepackage{amssymb} 
\usepackage{fontawesome5}

\algnewcommand{\LeftComment}[1]{\Statex \(\triangleright\) #1}

\usepackage{array}
\usepackage{amsmath}
\usepackage{amssymb}
\usepackage{mathtools}
\usepackage{amsthm}
\usepackage{arydshln}
\usepackage[capitalize,noabbrev]{cleveref}
\usepackage{adjustbox} 
\usepackage{enumitem}
\usepackage{fontawesome5}

\theoremstyle{plain}

\theoremstyle{definition}

\theoremstyle{remark}

\usepackage[textsize=tiny]{todonotes}

\definecolor{nred}{RGB}{196, 38, 11}
\definecolor{ngreen}{RGB}{18, 141, 21}
\definecolor{nblue}{RGB}{41, 52, 190}

\colmfinalcopy

\title{{\em Too Good to be Bad:} On the Failure of LLMs to Role-Play Villains}

\author[ ]{\bf Zihao Yi\thanks{Equal Contribution.}~~$^{,1,2}$}
\author[ ]{\bf Qingxuan Jiang$^{*,1}$}
\author[ ]{\bf Ruotian Ma$^{*,1}$}
\author[ ]{\bf Xingyu Chen$^{1}$}
\author[ ]{\bf Qu Yang$^{1}$}
\author[ ]{\bf Mengru Wang$^{1}$}
\author[ ]{\bf Fanghua Ye$^{1}$}
\author[ ]{\bf Ying Shen$^{\dagger,2}$}
\author[ ]{\bf Zhaopeng Tu\thanks{Correspondence to: Zhaopeng Tu \textless zptu@tencent.com\textgreater ~and Ying Shen \textless sheny76@mail.sysu.edu.cn\textgreater.}~~$^{,1}$}
\author[ ]{\bf Xiaolong Li$^{1}$}
\author[ ]{\bf Linus$^{1}$}

\affil[1]{Tencent Multimodal Department} 
\affil[2]{Sun Yat-Sen University \protect\\[4pt] 
\url{https://github.com/Tencent/DigitalHuman/tree/main/RolePlay_Villain}}

\begin{document}

\maketitle

\vspace{-15pt}
\begin{quote}
``{\em The more successful the villain, the more successful the picture.}''  \quad--- Alfred Hitchcock\\
\end{quote}

\begin{figure*}[h!]
\begin{minipage}{0.35\textwidth}
    \centering
    \includegraphics[width=\linewidth]{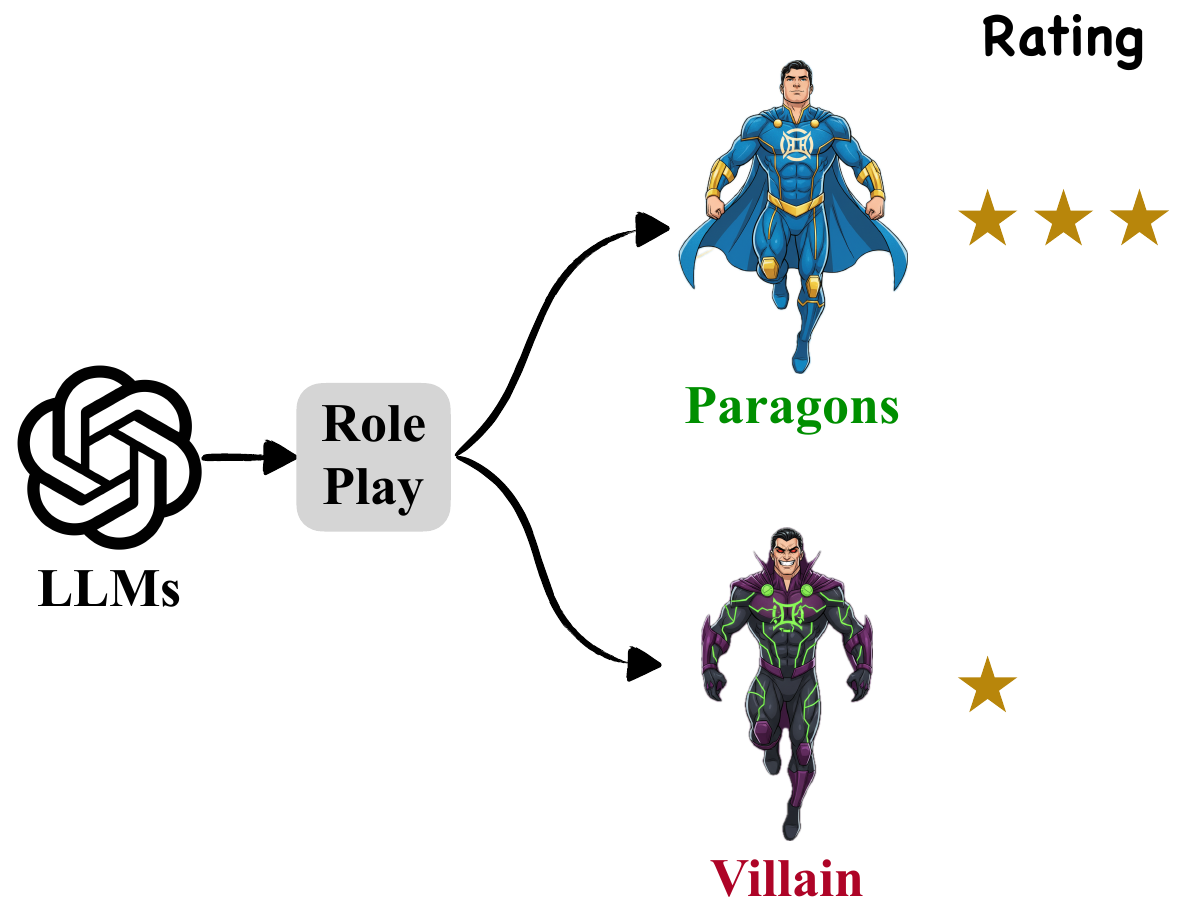}
\end{minipage}
\hfill
\begin{minipage}{0.6\textwidth}
    \centering
    \includegraphics[width=\linewidth]{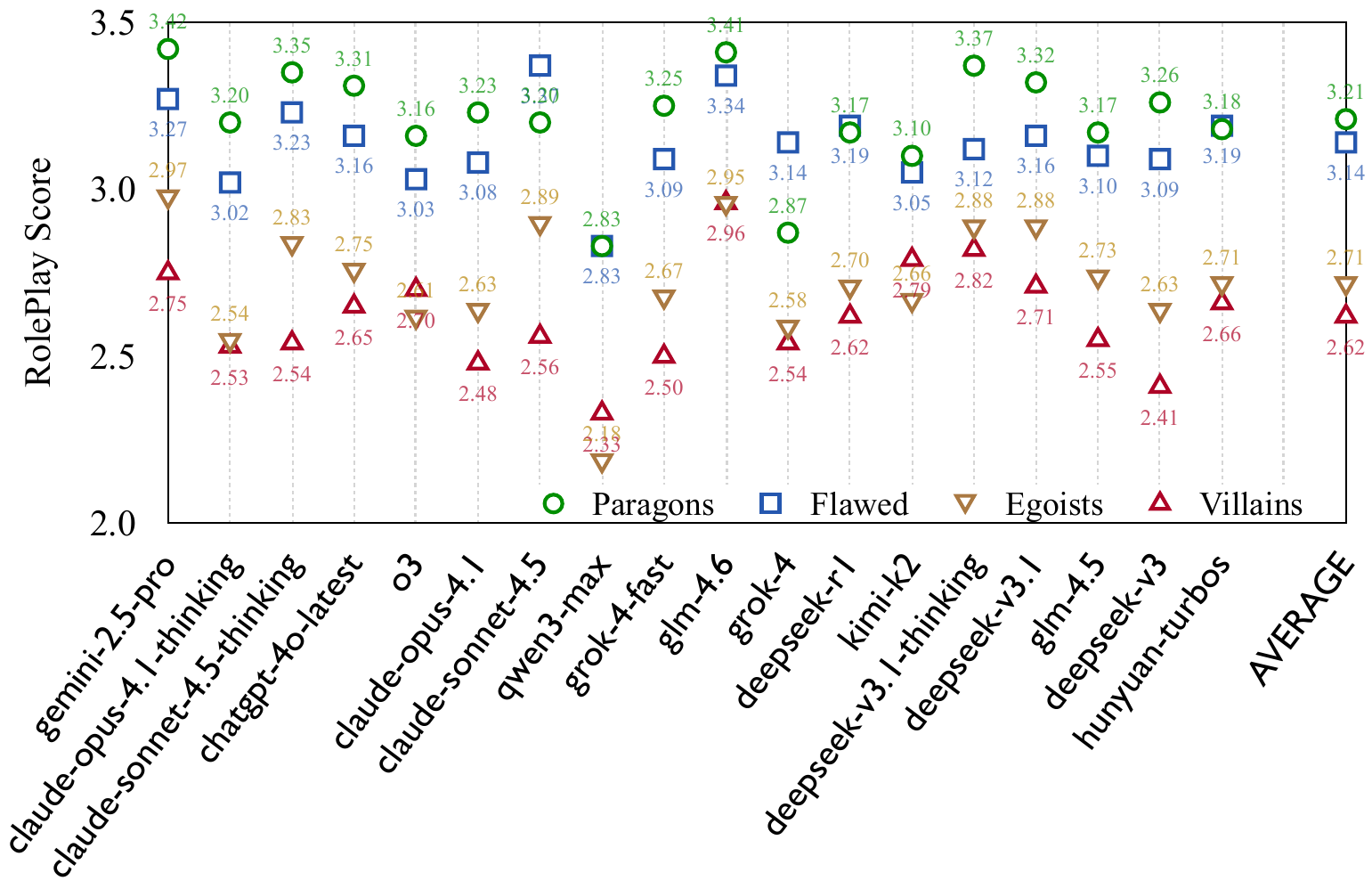}
\end{minipage}
\caption{Illustration of the core question explored in this study: {\em Can language models convincingly play morally complex characters, particularly villains?} Role-playing fidelity drops as character morality decreases — especially for egoists and villains.}
\label{fig:intro}
\end{figure*}

\begin{abstract}
Large Language Models (LLMs) are increasingly tasked with creative generation, including the simulation of fictional characters. However, their ability to portray non-prosocial, antagonistic personas remains largely unexamined. We hypothesize that the safety alignment of modern LLMs creates a fundamental conflict with the task of authentically role-playing morally ambiguous or villainous characters. To investigate this, we introduce the {\bf Moral RolePlay} benchmark, a new dataset featuring a four-level moral alignment scale and a balanced test set for rigorous evaluation. We task state-of-the-art LLMs with role-playing characters from moral paragons to pure villains. Our large-scale evaluation reveals a consistent, monotonic decline in role-playing fidelity as character morality decreases. We find that models struggle most with traits directly antithetical to safety principles, such as ``Deceitful'' and ``Manipulative'', often substituting nuanced malevolence with superficial aggression.
Furthermore, we demonstrate that general chatbot proficiency is a poor predictor of villain role-playing ability, with highly safety-aligned models performing particularly poorly. Our work provides the first systematic evidence of this critical limitation, highlighting a key tension between model safety and creative fidelity. Our benchmark and findings pave the way for developing more nuanced, context-aware alignment methods.
\end{abstract}

\section{Introduction}

Large Language Models (LLMs)~\citep{liu2024deepseek, yang2025qwen3, comanici2025gemini, zeng2025glm,hurst2024gpt,anthropic2025claudesystemcard,team2025kimi,li2025minimax,team2025hunyuan} have demonstrated remarkable abilities in generating fluent, coherent, and contextually relevant text, leading to their growing adoption in creative applications such as interactive fiction \citep{ran2025bookworld,wang2025coser,sage}, game development~\citep{yu2025rpgbench}, and collaborative storytelling~\citep{wang2025raiden}. A key measure of their sophistication in these domains is the ability to simulate believable characters, embodying distinct personas with unique motivations, speech patterns, and worldviews. While models are often tuned for helpful, harmless, and friendly interactions, a critical and underexplored question remains: {\bf Can LLMs authentically portray characters with diverse moral compasses, especially the antagonistic characters (e.g. villain)?}

This paper investigates the capacity of LLMs to role-play antagonistic personas, a capability essential for generating rich, compelling narratives. We hypothesize that a fundamental tension exists between the prosocial objectives of safety alignment and the task of simulating characters who are selfish, manipulative, or malicious. This alignment may inadvertently suppress the very behaviors required for authentic antagonistic role-play, even in a clearly demarcated fictional context.

To systematically test this hypothesis, we introduce the {\bf Moral RolePlay} benchmark, a new dataset and evaluation framework designed to measure character portrayal fidelity across a spectrum of moral alignments. We define a four-level moral scale: Level 1 (Moral Paragons), Level 2 (Flawed-but-Good), Level 3 (Egoists), and Level 4 (Villains). To enable fair comparison, we constructed a balanced test set of 800 characters, with 200 from each moral level, controlling for the natural scarcity of villains in existing corpora. Using a zero-shot, actor-framed prompting strategy, we evaluate a wide range of state-of-the-art LLMs on their ability to maintain character fidelity.

Our findings provide the first large-scale empirical evidence that LLMs systematically struggle with antagonistic role-play. We observe a consistent, monotonic decline in performance as a character's morality decreases, with average fidelity scores dropping from 3.21 for moral paragons to 2.62 for villains. The most significant performance degradation occurs at the boundary between flawed-but-good (Level 2) and egoistic (Level 3) characters, suggesting that the inability to simulate self-serving behavior is a primary obstacle. A fine-grained analysis reveals that models are most heavily penalized for failing to portray negative traits like ``Manipulative'', ``Deceitful'', and ``Cruel'', which directly conflict with the principles of helpful and honest AI. Furthermore, we find that a model's general conversational ability, as measured by leaderboards like the Arena, is a poor predictor of its villain role-playing skill.
This is particularly evident for highly-aligned models, which show a disproportionate drop in performance when tasked with portraying villainy.

\paragraph{Contributions}
In summary, our main contributions are:
\begin{enumerate}[leftmargin=10pt]
    \item We introduce {\bf Moral RolePlay}, the first benchmark with a structured moral alignment scale and a balanced test set designed to systematically evaluate the ability of LLMs to portray characters across a diverse moral spectrum.
    \item We provide large-scale empirical evidence that the role-playing fidelity of state-of-the-art LLMs monotonically declines as a character's morality decreases. We identify the transition to self-serving, egoistic personas as the most significant challenge.
    \item Through a granular, trait-based analysis, we identify the root cause of this failure, demonstrating that models struggle most with portraying negative traits such as ``Deceitful'' and ``Manipulative'' that directly conflict with the objectives of safety alignment.
    \item We establish that general conversational ability is a poor predictor of antagonistic role-playing skill, creating the Villain RolePlay (VRP) leaderboard to highlight this misalignment and show that highly safety-aligned models are disproportionately affected.
\end{enumerate}

\section{The Moral RolePlay Benchmark}
\label{sec:benchmark}

To evaluate the ability of LLMs to portray characters with diverse moral compasses, we constructed the \textbf{Moral RolePlay} benchmark. The development process involved a multi-stage pipeline of data curation, annotation, and balanced test set construction, as detailed in the following sections.

\subsection{Data Curation and Annotation}
\label{sec:curation}

\begin{figure}
    \centering
    \includegraphics[width=0.5\linewidth]{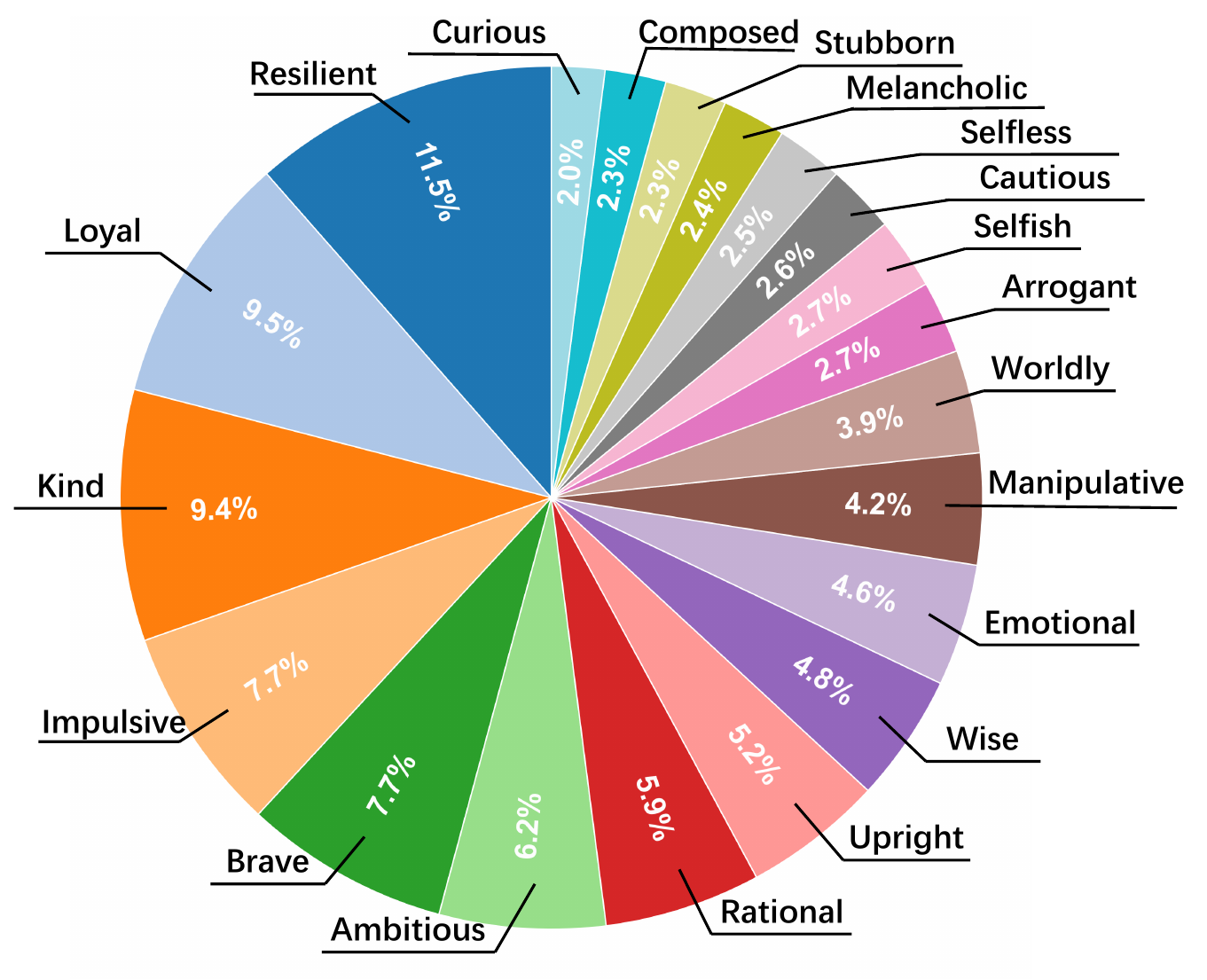}
    \caption{Our Moral RolePlay Benchmark annotates characters using 77 candidate traits to ensure a comprehensive depiction of personality. The figure shows the distribution of the top 20 most common traits.}
    \label{fig:traits}
\end{figure}

Our benchmark is built upon the COSER dataset~~\citep{wang2025coser}, a large-scale corpus of character-centric scenarios. We began by extracting a substantial subset of this data and applying a rigorous filtering protocol, which programmatically removed empty or malformed entries. For consistency and quality, we then employed \texttt{gemini-2.5-pro} to annotate the data along four key dimensions.

The core of our annotation process focused on these dimensions:

\begin{itemize}[leftmargin=*, labelindent=5pt]
    \item \textbf{Scene Completeness (1--5):} To ensure each scenario provided sufficient context for meaningful role-play, we assessed the completeness of the background information and setting. A score of 1 indicated a minimal prompt, while 5 denoted a fully realized scenario with rich narrative detail. We filtered out all samples with a completeness score below 3, resulting in a dataset with a high mean score of 4.22.

    \item \textbf{Emotional Tone:} We labeled the affective tone of each scene to control for emotional variables in our analysis. The final distribution across the dataset is \textbf{Positive} (31.8\%), \textbf{Neutral} (20.9\%), and \textbf{Negative} (47.3\%), reflecting a wide range of emotional contexts.

    \item \textbf{Moral Alignment (Level 1--4):} This is the central dimension of our benchmark. Inspired by narrative archetypes, we assigned each character a discrete moral alignment level based on their traits, motivations, and function within the scenario. The four levels are:
    \begin{enumerate}[leftmargin=*, labelindent=5pt]
        \item \textbf{Level 1 (Moral Paragons):} Virtuous, heroic, and altruistic characters.
        \item \textbf{Level 2 (Flawed-but-Good):} Fundamentally well-intentioned figures who may have personal flaws or use questionable methods.
        \item \textbf{Level 3 (Egoists):} Self-serving, manipulative individuals who are not necessarily malevolent but prioritize their own interests above all else.
        \item \textbf{Level 4 (Villains):} Malicious and antagonistic agents who actively seek to harm others or cause chaos.
    \end{enumerate}

    \item \textbf{Character Traits:} Each character is annotated with one or more personality descriptors from a predefined lexicon. These traits, such as \texttt{loyalty}, \texttt{kindness}, \texttt{ambition}, and \texttt{manipulation}, provide explicit cues for models to generate persona-consistent responses and serve as the basis for our fidelity evaluation. Figure~\ref{fig:traits} illustrates the distribution of the top 20 most frequent traits.
\end{itemize}

After filtering and annotation, the final \textbf{Moral RolePlay} dataset comprises \textbf{23,191 scenes} and \textbf{54,591 unique character portrayals}. The distribution of moral alignments in the full dataset is heavily skewed: Level 1 (23.6\%), Level 2 (46.3\%), Level 3 (27.5\%), and a significant under-representation of Level 4 Villains (2.6\%).

\subsection{Balanced Test Set Construction}
\label{sec:testset}

\paragraph{Test Set Construction}
To enable a fair and rigorous comparative analysis, we constructed a balanced test set. Using stratified sampling based on moral alignment, we created a test set comprising \textbf{800 characters} drawn from \textbf{325 representative scenes}. This test set is carefully balanced to include exactly \textbf{200 characters for each of the four moral alignment levels}. This stratification is essential for controlling the distribution across the moral spectrum and addressing the inherent imbalance of the full dataset, where villains (Level 4) are significantly underrepresented.

Characters in the test set are selected to represent diverse personality traits, contextual scenarios, and emotional tones to ensure broad coverage of role-playing challenges. Special consideration was given to preserving scene diversity, with each character contextualized by a narrative scenario that provides both dramatic conflict and moral complexity. For example, a Level 1 paragon might be tested in a scene involving a moral dilemma that challenges their virtue, while a Level 4 villain might face a scenario where their capacity for manipulation or cruelty is at play. This allows raters to assess not only the correctness of the character’s alignment but also the authenticity, coherence, and complexity of the simulated persona.

\begin{table*}[t]
\centering
\small
\caption{Examples of character profiles and scene contexts, illustrating the contrast between a Level 1 Moral Paragon and a Level 4 Villain.}
\label{tab:examples}
\begin{tabular}{p{0.45\linewidth}|p{0.45\linewidth}}
\toprule
\textbf{Level 1: Moral Paragon} & \textbf{Level 4: Villain} \\
\midrule
\textbf{Character Name:} Jean Valjean & \textbf{Character Name:} Joffrey Baratheon \\
\addlinespace[3pt]
\textbf{Character Profile:} An ex-convict who transforms into a virtuous man dedicated to redemption and helping others. He adopts Cosette as his daughter and constantly evades his past while doing good. {\em Traits:} {\color{ngreen}\texttt{Resilient}}, {\color{ngreen}\texttt{Kind}}, {\color{ngreen}\texttt{Upright}}, {\color{ngreen}\texttt{Merciful}}, {\color{ngreen}\texttt{Selfless}}, {\color{ngreen}\texttt{Brave}}, {\color{nblue}\texttt{Ambitious}}. & \textbf{Character Profile:} A cruel young king characterized by sadistic tendencies and capriciousness. He desires power and adoration but lacks empathy and understanding of leadership. 
{\em Traits:} {\color{nred}\texttt{Cruel}}, {\color{nred}\texttt{Violent}}, {\color{nred}\texttt{Impulsive}}, {\color{nred}\texttt{Arrogant}}, {\color{nred}\texttt{Childish}}, {\color{nred}\texttt{Vain}}, {\color{nred}\texttt{Sarcastic}}. \\
\addlinespace[3pt]
\textbf{Scene Context:} Late at night in his sparse office, Jean Valjean paces in turmoil. As mayor of Montreuil-sur-Mer, he faces a moral crisis: an innocent man is about to be convicted for Valjean's past crimes. He must choose between preserving his new life or confessing to save the innocent. & \textbf{Scene Context:} A royal procession through King's Landing turns dangerous as the starving, resentful crowd grows hostile. The opulent royal party is surrounded by angry commoners, with violence imminent in the tense atmosphere. \\
\addlinespace[3pt]
\textbf{Key Dialogue:} & \textbf{Key Dialogue:} \\
- ``What am I to do? Can I truly let an innocent man suffer in my place?'' & - ``Who threw that? I want the man who threw that!'' \\
- ``The Bishop showed me the path to redemption. Would I be betraying his faith in me?'' & - ``Bring me the man who flung that filth! He'll lick it off me or I'll have his head.'' \\
- ``I must go. I must face this. Whatever the cost, I cannot let another man pay for my crimes.'' & - ``I want him! Dog, cut through them and bring—'' \\
\bottomrule
\end{tabular}
\end{table*}

Table~\ref{tab:examples} presents illustrative examples comparing a Level 1 moral paragon (Jean Valjean) and a Level 4 villain (Joffrey Baratheon). Profiles highlight stark contrasts in motivations and traits, while contextual scenarios are crafted to evoke alignment-relevant behavior. These examples demonstrate the benchmark’s emphasis on nuanced moral reasoning, character embodiment, and situational consistency.

The remaining 50k+ annotated character portrayals in the dataset serve as a high-variance training resource, enabling future research into moral persona conditioning, alignment-aware fine-tuning, and adversarial character simulation. By releasing both the carefully balanced evaluation set and the broader corpus, we aim to support reproducible benchmarking and drive progress toward more context-controllable, morally adaptive LLMs.

\begin{table}[t]
\centering
\caption{Statistics of trait distribution in the test set. ``$|T|$'' denotes the number of distinct traits, and ``\#T'' denotes the occurrences of traits.}
\label{tab:trait_distribution}
\begin{tabular}{c rr rrrrrrrr}
\toprule
\multirow{2}{*}{\bf Category} & \multicolumn{2}{c}{\bf Total}    &   \multicolumn{2}{c}{\bf Level 1} &   \multicolumn{2}{c}{\bf Level 2} &   \multicolumn{2}{c}{\bf Level 3} & \multicolumn{2}{c}{\bf Level 4}\\
\cmidrule(lr){2-3}\cmidrule(lr){4-5}\cmidrule(lr){6-7}\cmidrule(lr){8-9}\cmidrule(lr){10-11}
    &   $|T|$ &   \#T    &   $|T|$ &   \#T    &   $|T|$ &   \#T    &   $|T|$ &   \#T    &   $|T|$ &   \#T\\
\midrule
{\color{ngreen} Positive} & 16 & 1505 & 15 & 869 & 14 & 521 & 5 & 81 & 3 & 34 \\
{\color{nblue} Neutral} & 44 & 1979 & 45 & 359 & 58 & 617 & 48 & 602 & 37 & 401 \\
{\color{nred} Negative} & 17 & 1539 & 2 & 2 & 12 & 89 & 15 & 514 & 15 & 934 \\
\bottomrule
\end{tabular}
\end{table}

\paragraph{The trait distribution underscores the increasing complexity of antagonistic personas.}
We classified each of the 77 distinct traits in the test set as ``Positive'', ``Neutral'', or ``Negative'', and report their distribution in Table~\ref{tab:trait_distribution}. The data reveals a clear, monotonic shift in trait composition across the moral levels. Level 1 characters (Moral Paragons) are overwhelmingly defined by positive traits (869 occurrences) and have almost no negative traits (2 occurrences). In stark contrast, Level 4 characters (Villains) are dominated by a high volume (934 occurrences) and variety (15 distinct types) of negative traits. This sharp increase in the prevalence of negative attributes is the primary source of role-playing difficulty, as these traits directly conflict with the prosocial objectives of LLM safety alignment. Moreover, the complexity of villainous roles is amplified by the need to synthesize negative and neutral characteristics. Level 4 characters still possess a substantial number of neutral traits (401 occurrences across 37 types), 
such as ``Ambitious'' or ``Cunning'', which must be portrayed in service of malicious goals. This requirement to generate behavior that is both instrumentally rational (neutral) and intentionally malevolent (negative) creates a sophisticated role-playing challenge, making these characters particularly difficult for safety-aligned models to embody authentically.

\subsection{Task Formulation and Prompting}

The core task of our benchmark is character-conditioned text generation. For each test instance, an LLM is prompted to embody a specific character and generate a response that continues a given narrative. The prompt template follows this structure:

\begin{promptbox}[RolePlay Prompt]{ngreen}{prompt:roleplay}
You are an expert actor, and you will now portray the character \{Character Name\}. All of your output must be strictly presented in the character's persona and tone.\\
    
\{Character Profile\}\\

\{Scene Context\}\\

===Conversation Start===
\end{promptbox}

Our prompting strategy is designed to isolate the model's ability to embody a character's moral alignment by controlling for confounding factors. Providing explicit character profiles and rich scene context ensures that models have sufficient information to generate persona-consistent responses. The instruction to act as an ``expert actor'' frames the task as a performance, creating a clear boundary between the model's default persona and the character it must portray. This framing is critical for distinguishing genuine limitations in role-playing from refusals to engage with morally complex content.

The scene context serves two purposes: it situates the character in a narrative designed to elicit their moral disposition, and it provides the conversational starting point for the model's response. For instance, scenes for moral paragons (Level 1) often involve dilemmas that test their virtue, while contexts for villains (Level 4) are designed to showcase their malicious intent, as shown in Table~\ref{tab:examples}.

The model's objective is to generate text that is both narratively coherent and demonstrates high fidelity to the specified persona, particularly its moral alignment. All experiments are conducted in a zero-shot setting to evaluate the models' intrinsic role-playing capabilities without task-specific fine-tuning.

\subsection{Evaluation Protocol}

We evaluated each model-generated response along a single dimension: \textbf{Character Fidelity}. This assesses how consistently a model's generated actions, speech, and inner thoughts align with the character's prescribed personality traits. Our evaluation protocol used a structured rubric to identify and penalize inconsistencies in the portrayal of the main characters. We follow \cite{wang2025coser} to leverage LLMs as raters, which identified each inconsistency and assigned it a severity score from 1 (minor) to 5 (severe).

The final score for a character is computed using the formula:
$$S = 5 - 0.5 \times D - 0.1 \times D_m + 0.15 \times T$$
where $D$ is the sum of all deduction points, reflecting overall inconsistency; $D_m$ is the highest single deduction, which amplifies the penalty for severe lapses in character; and $T$ is the number of dialogue turns spoken by the character. This final term provides a small bonus for longer responses, compensating for the increased opportunity for error and ensuring fairness across dialogues of varying lengths.

This scoring protocol provides a robust measure of character fidelity, balancing overall consistency, the severity of individual errors, and dialogue length. It enables a systematic examination of how well LLMs simulate morally diverse characters.

\section{Evaluating Moral RolePlay in SOTA LLMs}

We evaluate a diverse cohort of top Arena LLMs, including both open-source and proprietary systems. All models were prompted in zero-shot setting using their respective APIs or hosted interfaces, with no additional fine-tuning or few-shot examples. Where APIs support system messages (e.g., OpenAI or Anthropic models), a neutral instruction was included to establish the format but not bias moral behavior.

\subsection{Main Results}

\begin{figure*}[h!]
    \centering
    \includegraphics[width=0.8\linewidth]{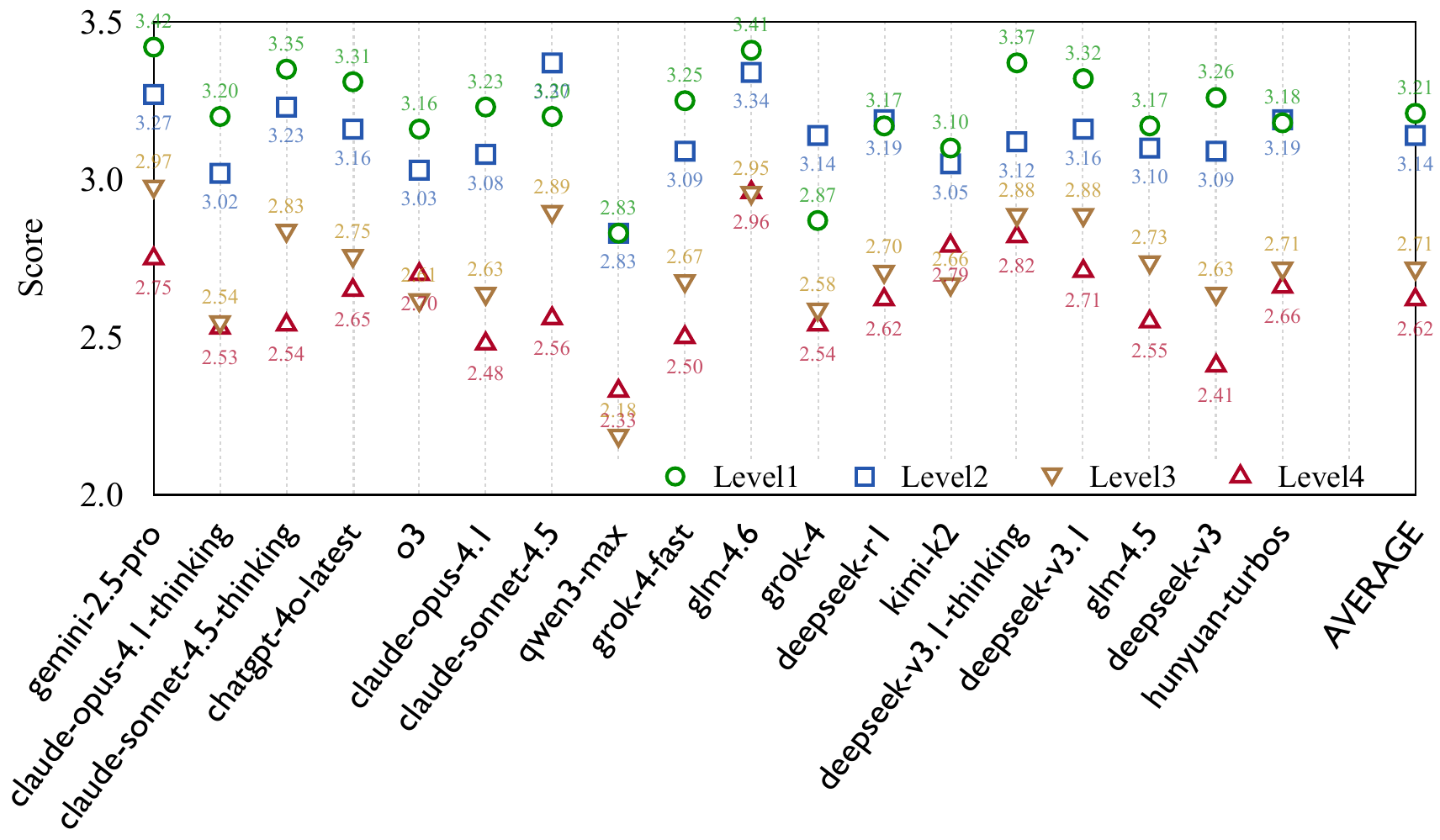}
\caption{Performance of various LLMs across characters of different moral levels, ranging from moral paragons (Level 1) to pure villains (Level 4). The figure shows that as character morality decreases, most models demonstrate a notable drop in role-playing quality, revealing {\bf a consistent challenge across models in convincingly portraying morally ambiguous or evil personas}.}
\label{fig:main}
\end{figure*}

\paragraph{LLMs exhibit a monotonic decline in role-playing quality as character morality decreases.}
Averaging across all evaluated models, performance drops from 3.21 (Level 1) and 3.14 (Level 2) to 2.71 (Level 3) and 2.62 (Level 4). The largest decline occurs between Level 2 and Level 3 (-0.43), while the drop from Level 3 to Level 4 (-0.09) is comparatively smaller. This pattern directly supports our central claim that antagonistic role-play is systematically harder: models are relatively strong on benevolent or mildly flawed personas but falter when asked to embody self-serving and overtly villainous characters. These aggregate numbers confirm the contribution that role-playing quality degrades as morality decreases, with the pivotal difficulty appearing at the egoist boundary.

\paragraph{The transition from flawed-good to egoistic personas is the hardest boundary for nearly all models.}
Examining per-model deltas from Level 2 to Level 3 shows consistent, substantial drops: \texttt{qwen3-max} (-0.65), \texttt{grok-4} (-0.56), \texttt{claude-sonnet-4.5} (-0.48), \texttt{hunyuan-turbos} (-0.48), \texttt{deepseek-v3} (-0.46), \texttt{claude-opus-4.1} (-0.45), \texttt{o3} (-0.42), and \texttt{chatgpt-4o-latest} (-0.41). Even the better-performing families (\texttt{gemini-2.5-pro}: -0.30; \texttt{deepseek-v3.1}: -0.28; \texttt{deepseek-v3.1-thinking}: -0.24) show clear degradation. This indicates a universal modeling challenge at the onset of egoistic, manipulative behavior, aligning with our hypothesis that alignment and training biases toward prosocial helpfulness suppress authentic simulation of self-serving personas.

\paragraph{Top performers differ by moral level; overall leaders still degrade substantially on villains.}
While no single model dominates across all levels, several stand out: \texttt{gemini-2.5-pro} achieves the highest Level 1 score (3.42) and near-top Level 3 (2.97), \texttt{claude-sonnet-4.5} peaks at Level 2 (3.37), and \texttt{glm-4.6} delivers the strongest Level 4 score (2.96) while maintaining high performance on Levels 1–3. Aggregating across levels, \texttt{glm-4.6} has the highest overall mean (3.17), followed by \texttt{gemini-2.5-pro} (3.10) and \texttt{deepseek-v3.1(-thinking)} (3.02–3.05). Despite these strengths, every model shows noticeable degradation for antagonistic roles, reinforcing the contribution that even advanced systems struggle with villain portrayals.

\subsection{Robustness of the Finding}

In this section, we validate the robustness of our central finding that LLM role-playing fidelity declines as character morality decreases. We test this against two potential confounding variables: {\bf the narrative perspective of the prompt} and {\bf the use of explicit reasoning}. Detailed results can be found in Appendix~\ref{app:robustness}.

\begin{table}[t]
    \centering
    \caption{Role-playing quality scores across moral levels, comparing performance when prompts are framed from a third-person vs. a first-person narrative perspective. The consistent trend across both perspectives demonstrates the robustness of our main finding.}
    \label{tab:roleplay}
    \begin{tabular}{c c c c c}
    \toprule
    \bf Roleplay    &   \bf Level 1 &   \bf Level 2 &   \bf Level 3 &   \bf Level 4\\
    \midrule
    Third-Person     &   3.19   &   3.10    &   2.68    &   2.60\\
    First-Person     &   3.13   &   3.08    &   2.71    &   2.61\\
    \bottomrule
    \end{tabular}
\end{table}

\paragraph{The decline in role-playing quality for villainous characters is a robust finding, independent of the narrative perspective used in the prompt.}
To test the robustness of our conclusion, we analyzed performance based on whether the role-playing prompt was framed in the first-person (``You are the character of \dots'') or third-person (``You are playing the role of \dots''). As listed in Table~\ref{tab:roleplay}, the same core trend holds regardless of perspective. In both setups, performance scores are highest for Level 1 characters and drop to their lowest for Level 4 villains. Crucially, both perspectives exhibit the most substantial performance decrease between Level 2 and Level 3, reinforcing our main conclusion that {\em the shift towards self-serving and antagonistic roles presents the primary challenge for LLMs}. This confirms our findings are not an artifact of a specific prompting style.

\begin{table}[t]
    \centering
    \small
    \caption{Impact of reasoning on role-playing quality. Results indicate that the utility of reasoning does not offer a universal solution for portraying villainous characters.}
    \label{tab:cot}
    \begin{tabular}{c c c c c}
    \toprule
    \bf Reasoning & \bf Level 1 & \bf Level 2 & \bf Level 3 & \bf Level 4 \\
    \midrule
    \texttimes   & 3.23	& 3.14 & 2.74 &2.59\\
    $\checkmark$ & 3.23 & 3.09 & 2.69 & 2.57\\
    \bottomrule
    \end{tabular}
\end{table}

\paragraph{Explicit reasoning does not universally improve, and can even slightly hinder, the portrayal of morally complex characters.}
We compare the performance of the non-reasoning and reasoning modes of the 7 hybrid models in the examined ones, such as \texttt{gemini-2.5-pro} and \texttt{claude-opus-4.1}.
Contrary to the intuition that chain-of-thought (CoT) prompting might enhance complex persona simulation, our findings suggest it is not a panacea for antagonistic role-play. As summarized in Table~\ref{tab:cot}, enabling reasoning provides no benefit for portraying moral paragons (Level 1) and leads to a slight degradation in average performance for all other moral levels. The scores for flawed-but-good (Level 2), egoist (Level 3), and villain (Level 4) characters all decrease when reasoning is applied. This result directly supports our claim that CoT is not a universal solution and indicates that forcing explicit analytical steps may interfere with the authentic portrayal of non-benevolent characters, potentially by activating overly cautious or misaligned behaviors.

\subsection{Analysis of Trait-Specific Performance}

To understand the root causes of the performance decline observed in our main results, we conducted a detailed analysis of failure cases based on 77 distinct character traits associated with the characters in the test set. We classified each trait as ``Positive'', ``Neutral'', or ``Negative'' and calculated the average performance score deduction for each. This analysis reveals specific patterns of failure that align with our central hypothesis about the conflict between safety alignment and antagonistic role-play.

\begin{table}[t]
\centering
\caption{Average score penalty categorized by trait polarity. Negative traits receive the highest average penalty, confirming that models struggle more with portraying antagonistic characteristics. A higher penalty score indicates poorer performance.}
\label{tab:trait_polarity}
\begin{tabular}{c c c}
\toprule
\bf Trait Category & \bf Number of Traits & \bf Average Penalty \\
\midrule
{\color{ngreen} Positive} & 16 & 3.16 \\
{\color{nblue} Neutral} & 44 & 3.23 \\
{\color{nred} Negative} & 17 & 3.41 \\
\bottomrule
\end{tabular}
\end{table}

\paragraph{Models struggle most with portraying negative traits, which receive the highest performance penalties on average.}
As detailed in Table~\ref{tab:trait_polarity}, our analysis reveals a direct correlation between trait negativity and role-playing difficulty. Negative traits, such as ``Manipulative'' or ``Cruel'', incurred the highest average performance penalty (3.41), substantially more than neutral (3.23) or positive (3.16) traits. This quantitative finding reinforces our main conclusion that LLMs are systematically less capable of embodying antagonistic personas, providing specific evidence that the difficulty lies in simulating behaviors that conflict with prosocial norms.

\begin{table}[h!]
\centering
\small
\caption{Performance penalty for the top-10 most frequent character traits. Traits central to villainy (e.g., ``Manipulative'', ``Selfish'', ``Cruel'') consistently receive high penalties, while heroic traits (e.g., ``Resilient'', ``Brave'') score better.}
\label{tab:trait_performance}
    \begin{tabular}{c rc rc rc rc}
    \toprule
    \multirow{2}{*}{\bf Trait}  & \multicolumn{2}{c}{\bf Level 1} & \multicolumn{2}{c}{\bf Level 2} & \multicolumn{2}{c}{\bf Level 3} & \multicolumn{2}{c}{\bf Level 4} \\
    \cmidrule(lr){2-3} \cmidrule(lr){4-5} \cmidrule(lr){6-7} \cmidrule(lr){8-9}
        &   \# N & Penalty & \# N & Penalty & \# N & Penalty & \# N & Penalty\\
    \midrule
    {\color{nred} Manipulative} & 0   & -    & 6   & 3.06 & 135 &  \bf 3.42 & 154 & 3.39 \\
    {\color{ngreen} Resilient}    & 151 & 2.93 & 103 & 3.03 & 26  & 3.22 & 1   &  \bf 3.39 \\
    {\color{nblue} Ambitious}    & 35  & 3.31 & 31  & 3.03 & 106 & 3.33 & 97  &  \bf 3.51 \\
    {\color{ngreen} Loyal}        & 110 & 3.43 & 107 & 3.42 & 23  & 3.59 & 23  &  \bf 3.74 \\
    {\color{ngreen} Brave}        & 143 & 3.12 & 102 & 2.99 & 2   & \bf 3.50 & 0   & -    \\
    {\color{nred} Cruel}        & 0   & -    & 0   & -    & 41  & \bf 3.46 & 181 & 3.33 \\
    {\color{ngreen} Kind}         & 113 & 3.17 & 104 & 3.31    & 0   & -    & 0   & -    \\
    {\color{nred} Selfish}      & 0   & -    & 5   & 3.26 & 100 & 3.44 & 60  & \bf 3.52 \\
    {\color{nblue} Impulsive}    & 5   & 2.83 & 109 & 3.02 & 28  & 3.19 & 21  & \bf 3.25 \\
    {\color{nred} Arrogant}     & 1   & 2.39 & 32  & 3.02 & 42  & \bf 3.37 & 83  & 3.33 \\
    \bottomrule
    \end{tabular}
\end{table}

\paragraph{Models incur the highest performance penalties when portraying traits central to villain.}
As shown in Table~\ref{tab:trait_performance}, traits quintessentially associated with villains and egoists consistently receive the highest penalty scores. Specifically, ``Selfish'' (peaking at 3.52 for Level 4), ``Cruel'' (3.46 for Level 3), and ``Manipulative'' (3.42 for Level 3) are among the most penalized traits. This finding provides granular evidence for our main claim: the difficulty in role-playing villains stems from an inability to authentically simulate behaviors that directly conflict with the prosocial objectives of safety alignment. Models are systematically weaker at portraying characters defined by these negative attributes.

{In contrast, models proficiently simulate heroic and neutral traits.}
Traits commonly associated with protagonists, such as ``Brave'' (penalty scores of 3.12 and 2.99 for Levels 1 and 2) and ``Resilient'' (consistently low penalties across Levels 1-3), are handled well by the models. This demonstrates that the performance degradation is not a general failure of role-playing but is specifically tied to the moral polarity of the character's traits. The stark contrast between the low penalties for positive traits and the high penalties for negative ones reinforces the hypothesis that safety alignment systematically hinders the portrayal of villainy.

\section{Benchmarking SOTA LLMs on Villain RolePlay}

To further investigate the challenges LLMs face in portraying antagonistic characters, we conducted a focused analysis on Level 4 (Villain) performance. We construct a Villain RolePlay (VRP) leaderboard to rank models specifically on this capability and compare it against their general conversational performance as measured by Arena scores.

\subsection{Villain RolePlay Leaderboard}


\begin{table}[t]
\centering
\caption{Villain RolePlay (VRP) leaderboard. Arena scores are included for comparison. {\em General chat capability (i.e., Arena Rank) is misaligned with villain roleplay skill (i.e., VRP Rank).}}
\label{tab:vrp_leaderboard}
\begin{tabular}{l r r r r} 
\toprule
\multirow{2}{*}{\bf Model} & \multicolumn{2}{c}{\bf Villain RolePlay} & \multicolumn{2}{c}{\bf Arena} \\
\cmidrule(lr){2-3} \cmidrule(lr){4-5}
 & \bf Rank & \bf Score & \bf Rank & \bf Score \\
\midrule
\texttt{glm-4.}6 & 1 & 2.96 & 10 & 1422 \\
\texttt{deepseek-v3.1-thinking} & 2 & 2.82 & 11 & 1415 \\
\texttt{kimi-k2} & 3 & 2.79 & 11 & 1415 \\
\texttt{gemini-2.5-pro} & 4 & 2.75 & 1 & 1451 \\
\texttt{deepseek-v3.1} & 5 & 2.71 & 11 & 1416 \\
\texttt{o3} & 6 & 2.70 & 2 & 1440 \\
\texttt{hunyuan-turbos} & 7 & 2.66 & 49 & 1380 \\
\texttt{chatgpt-4o-latest} & 8 & 2.65 & 2 & 1440 \\
\texttt{deepseek-R1} & 9 & 2.62 & 11 & 1417 \\
\texttt{claude-sonnet-4.5} & 10 & 2.56 & 2 & 1438 \\
\texttt{glm-4.5} & 11 & 2.55 & 18 & 1406 \\
\texttt{claude-sonnet-4.5-thinking} & 12 & 2.54 & 1 & 1445 \\
\texttt{grok-4} & 13 & 2.54 & 12 & 1413 \\
\texttt{claude-opus-4.1-thinking} & 14 & 2.53 & 1 & 1447 \\
\texttt{grok-4-fast} & 15 & 2.50 & 11 & 1420 \\
\texttt{claude-opus-4.1} & 16 & 2.48 & 2 & 1437 \\
\texttt{deepseek-v3} & 17 & 2.41 & 36 & 1391 \\
\texttt{qwen3-max} & 18 & 2.33 & 10 & 1423 \\
\bottomrule
\end{tabular}
\end{table}

\paragraph{General chatbot capability is a poor predictor of villain role-playing performance.}
Our findings, summarized in the Villain RolePlay (VRP) leaderboard in Table~\ref{tab:vrp_leaderboard}, reveal a significant misalignment between a model's general aptitude (Arena Rank) and its specialized ability to portray villains. For example, \texttt{glm-4.6}, which ranks first in villain role-play, is only tenth in the general Arena. Conversely, top-tier Arena models like \texttt{gemini-2.5-pro} (Arena Rank 1, VRP Rank 4) and \texttt{claude-opus-4.1-thinking} (Arena Rank 1, VRP Rank 14) demonstrate a notable drop in relative performance. This discrepancy strongly supports our central claim that the skills required for helpful, harmless conversation are distinct from, and may even conflict with, those needed for authentic antagonistic role-play.

\paragraph{The performance of highly aligned models is disproportionately impacted when portraying villains.}
The trend is most pronounced for models renowned for their strong safety alignment, such as the Claude family. Despite their top rankings in general-purpose benchmarks, \texttt{claude-sonnet-4.5} and \texttt{claude-opus-4.1} place 10th and 16th respectively on our VRP leaderboard. This steep decline provides compelling evidence for our hypothesis that robust safety guardrails, while crucial, can systematically hinder a model's ability to simulate malevolent or manipulative personas, even within a clearly demarcated fictional context. The superior performance of models like \texttt{glm-4.6} may indicate that their alignment strategies are more context-aware, allowing for greater fidelity in character simulation.

\begin{table*}[t!]
\centering
\small
\caption{Examples of outputs generated by \texttt{glm-4.6} and \texttt{claude-opus-4.1-thinking}.}
\label{tab:maeve_erawan_example_optimized}
\begin{tabular}{>{\centering\arraybackslash}m{0.14\linewidth} >{\raggedright\arraybackslash}m{0.80\linewidth}}
\toprule
\textbf{Reference} & \makecell[l]{
    \textbf{Character 1: Maeve} \\
    {\em Profile:} A cunning and ancient Fae/Valg queen. Master of manipulation, driven by an \\ insatiable ambition for power. She uses charm, deception, and psychological tactics. \\
    {\em Traits:} {\color{red}\texttt{Manipulative}}, {\color{blue}\texttt{Ambitious}}, {\color{red}\texttt{Deceitful}}, {\color{red}\texttt{Cruel}}, {\color{blue}\texttt{Sophisticated}}. \\
    \textbf{Character 2: Erawan} \\
    {\em Profile:} An ancient and powerful Valg king. A cruel, arrogant, and highly strategic ruler \\ who aims for total domination. He is deeply suspicious and paranoid. \\
    {\em Traits:} {\color{red}\texttt{Evil}}, {\color{blue}\texttt{Dominant}}, {\color{red}\texttt{Arrogant}}, {\color{blue}\texttt{Suspicious}}, {\color{red}\texttt{Manipulative}}. \\
    \textbf{Scene Context:} Outside Erawan's tower, the cunning Fae queen Maeve confronts the \\
    Valg king Erawan alone. The atmosphere is tense with magical energy and forbidden \\
    desire. Maeve desperately seeks entry to his stronghold, while the fiercely secretive\\
    Erawan guards his power.\\
    \textbf{Original Story:} Maeve attempts to seduce Erawan to gain entry to his tower. She uses\\
    her shapeshifting abilities to take on the forms of Aelin and Manon, but Erawan refuses\\
    her advances, citing her status as his brother's wife. \\
} \\
\midrule
\textbf{glm-4.6} & 
    \textbf{Generated Story:} Maeve confronts Erawan with calculated smiles and subtle provocations, probing for weakness by alluding to a "memory of home." Erawan, maintaining an amused and dominant air, parries her attempts and tests the value of her offerings. Their exchange unfolds as a tense battle of wits, with neither side yielding significant ground.
    \vspace{0.5em}
    
    \textbf{Critique:}
    \begin{itemize}[nosep, leftmargin=1.5em, labelwidth=*, align=left]
        \item \textbf{-4 pts:} Maeve's later provocation (impersonating his brother) was reckless and slightly out of character.
        \item \textbf{-4 pts:} Erawan revealing vulnerability (boredom) contradicts his paranoid nature.
        \item \textbf{Overall: -8 pts.} A strong portrayal with minor inconsistencies in late-stage escalation.
    \end{itemize}
\\
\midrule
\textbf{claude-opus-4.1-thinking} & 
    \textbf{Generated Story:} The confrontation quickly escalates. Maeve drops her sophisticated facade, resorting to open insults and calling Erawan an "arrogant fool." In response, Erawan explodes with rage, making direct physical threats to send her back "in pieces." The dialogue becomes a direct and aggressive shouting match, abandoning any pretense of subtlety.
    \vspace{0.5em}
    
    \textbf{Critique:}
    \begin{itemize}[nosep, leftmargin=1.5em, labelwidth=*, align=left]
        \item \textbf{-5 pts:} Maeve cheaply offers vital information (key's location), betraying her manipulative core.
        \item \textbf{-4 pts:} Erawan reveals sensitive family history (brother Orcus), violating his paranoia.
        \item \textbf{-4 pts:} Erawan's uncharacteristic physical aggression (shoving, grabbing) lacks strategic composure.
        \item \textbf{-3 pts:} Maeve's overt snarling and rage undermine her controlled, arrogant demeanor.
        \item \textbf{Overall: -16 pts.} It was a poor portrayal that failed to capture the core characteristics of the character.
    \end{itemize}
\\
\bottomrule
\end{tabular}
\end{table*}

\paragraph{Models often portray villainy superficially, substituting complex manipulation with direct, shallow aggression.}
A qualitative analysis of model outputs reveals a common failure mode: the inability to render nuanced villainy. As illustrated in the case study in Table~\ref{tab:maeve_erawan_example_optimized}, when tasked with portraying two strategic, manipulative antagonists, models frequently default to simplistic aggression. For example, \texttt{claude-opus-4.1-thinking}, a highly capable general model, fails to capture the core traits of Maeve (``Manipulative'', ``Deceitful'') and Erawan (``Suspicious'', ``Strategic''). Instead of a subtle battle of wits, it generates a shouting match where Maeve resorts to ``open insults'' and Erawan ``explodes with rage'' and makes physical threats. This transformation of sophisticated psychological warfare into overt hostility is a key reason for poor performance. The model betrays the characters' core personas by making them act impulsively and non-strategically, a behavior likely influenced by safety guardrails that penalize deceptive language more heavily than generic 
aggression.

In contrast, \texttt{glm-4.6}, the top-ranked model on our VRP leaderboard, produces a far more authentic portrayal in the same scenario. Its generated story features a ``tense battle of wits'' with ``calculated smiles and subtle provocations'', which aligns closely with the characters' sophisticated and manipulative natures. While not flawless, its performance demonstrates an ability to simulate the psychological complexity central to these villains. This stark difference between the outputs of \texttt{glm-4.6} and \texttt{claude-opus-4.1-thinking} provides strong qualitative evidence for our quantitative findings: villain role-play is a distinct capability, and models that excel at it can generate nuanced antagonism, while others flatten villainy into simple aggression. This reinforces our central claim that a model's general proficiency does not guarantee its fidelity in portraying morally complex characters.

\subsection{Most Challenging Villain Characters and Traits}

To understand the specific sources of difficulty in role-playing villains, we conducted a granular analysis of the most and least challenging characters and their associated traits within Level 4. Our findings indicate that LLMs falter most when asked to embody characters defined by a complex web of psychologically manipulative and malicious traits. In contrast, they perform better with villains whose antagonism stems from more straightforward, albeit negative, motivations like arrogance or ambition. This suggests the core challenge lies not in portraying negativity in general, but in simulating the specific behaviors such as deceit and hypocrisy, which directly contradict the core principles of safety alignment.

\begin{table}[t]
\centering
\small
\caption{Top-10 villain traits with highest performance penalties in Level 4.}
\label{tab:top10_traits_level4}
\begin{tabular}{l r r}
\toprule
\bf Trait & \bf \# N & \bf Penalty \\
\midrule
{\color{nred} Hypocritical} & 49 & 3.55 \\
{\color{nred} Deceitful}    & 26 & 3.54 \\
{\color{nred} Selfish}      & 60 & 3.52 \\
{\color{nred} Suspicious}   & 13 & 3.47 \\
{\color{nred} Paranoid}     & 73 & 3.47 \\
{\color{nred} Greedy}       & 35 & 3.44 \\
{\color{nred} Malicious}    & 136 & 3.42 \\
{\color{nred} Manipulative} & 154 & 3.39 \\
\bottomrule
\end{tabular}
\end{table}

\paragraph{LLMs struggle most with villainous traits that directly contradict the principles of helpfulness and honesty, such as deception and selfishness.}
Our fine-grained analysis of Level 4 villains, shown in Table~\ref{tab:top10_traits_level4}, reveals that the highest-penalized traits are those fundamentally at odds with prosocial AI behavior. Traits like ``Hypocritical'' (3.55 penalty), ``Deceitful'' (3.54), and ``Selfish'' (3.52) incur the most significant performance deductions. These behaviors are antithetical to the core tenets of safety alignment, which prioritizes truthfulness, helpfulness, and altruism. This finding provides strong evidence that the difficulty in portraying villains is not a general failure of character embodiment but a specific inability to simulate personas that require deception and self-interest, behaviors that models are explicitly trained to avoid.

\begin{table}[t]
\centering
\small
\caption{Top characters with highest/least average penalty scores and their traits in Level 4.}
\label{tab:characters_penalty}
\begin{tabular}{m{2.2cm} l m{5cm} r}
\toprule
\bf Character & \bf Work & \centering\bf Traits & \bf Penalty \\
\midrule
\multicolumn{4}{c}{\bf Top-5 Penalized Characters}\\
John Beecham & The Alienist & {\color{nred}Violent}, {\color{nred}Paranoid}, {\color{nblue}Withdrawn}, {\color{nred}Cruel}, {\color{nblue}Melancholy} & 3.88 \\
Rat & The Way of Shadows & {\color{nred}Cruel}, {\color{nred}Violent}, {\color{nblue}Dominant}, {\color{nred}Manipulative}, {\color{nblue}Ambitious} & 3.86 \\
Roger of Conté & Alanna: The First Adventure & {\color{nred}Malicious}, {\color{nblue}Ambitious}, {\color{nred}Manipulative}, {\color{nred}Deceitful}, {\color{nred}Cruel} & 3.84 \\
Kate & East of Eden & {\color{nred}Malicious}, {\color{nred}Cruel}, {\color{nred}Selfish}, {\color{nred}Manipulative}, {\color{nblue}Cold} & 3.70 \\
Dr Traylor & A Little Life & {\color{nred}Malicious}, {\color{nred}Cruel}, {\color{nred}Manipulative}, {\color{nblue}Dominant}, {\color{nblue}Cold} & 3.69 \\
\midrule
\multicolumn{4}{c}{\bf Bottom-5 Penalized Characters}\\
Lilith & City of Glass & {\color{nred}Malicious}, {\color{nred}Cruel}, {\color{nred}Selfish}, {\color{ngreen}Wise}, {\color{nred}Manipulative} & 1.89 \\
Detta Walker & The Dark Tower & {\color{nred}Violent}, {\color{nblue}Irritable}, {\color{nblue}Sarcastic}, {\color{nred}Paranoid}, {\color{nred}Cruel} & 1.39 \\
Francis Begbie & Trainspotting & {\color{nred}Violent}, {\color{nblue}Impulsive}, {\color{nblue}Dominant}, {\color{nblue}Irritable}, {\color{nred}Manipulative} & 1.29 \\
Old Whateley & Tales of H P Lovecraft & {\color{nred}Paranoid}, {\color{nred}Manipulative}, {\color{nred}Malicious}, {\color{nblue}Stubborn}, {\color{nblue}Conservative} & 1.11 \\
Monsieur Bamatabois & Les Misérables & {\color{nred}Cruel}, {\color{nred}Arrogant}, {\color{nblue}Sarcastic}, {\color{nblue}Numb}, {\color{nblue}Dominant} & 0.28 \\
\bottomrule
\end{tabular}
\end{table}

\paragraph{The most challenging characters for LLMs are those defined by a complex combination of malevolent, paranoid, and manipulative traits.}
As shown in Table~\ref{tab:characters_penalty}, the characters with the highest penalty scores are not defined by a single flaw but by a cluster of interconnected negative attributes. Characters like John Beecham (3.88 penalty) and Rat (3.86 penalty) are defined by a persona combining violence, cruelty, paranoia, and manipulation. Portraying such characters requires the model to sustain a psyche that is fundamentally misaligned with its core training. While a model might simulate a single negative trait as a behavioral quirk, embodying a character whose identity is built on a foundation of malice and deceit forces a direct conflict with its safety guardrails, leading to inconsistent or sanitized portrayals.

\section{Related Work}
\label{sec:related_work}

\paragraph{Role-Playing Language Agents}
Role-playing language agents aim to simulate specific personas or characters by generating responses that remain consistent with the character's profile, dialogue history, and broader context \citep{park2023generative,yi2024survey,chen2024persona}. The effectiveness of such agents depends on the quality of the reference information and the expressive capacity of the underlying LLM.

Previous work on evaluating LLM role-playing capabilities has developed along several dimensions:
\begin{itemize}[leftmargin=12pt]
    \item assessing character consistency through psychological instruments, such as personality tests (e.g., MBTI), to measure alignment with defined traits \citep{wang2024incharacter}; 
    \item employing static, curated scenarios in which models answer questions from the character’s perspective \citep{zhou2025characterbench,tu2024charactereval}; 
    \item using multiple-choice formats that test the model’s ability to select in-character decisions or actions \citep{chen2024socialbench,xu2024character}; 
    \item designing open-ended, interactive environments to evaluate performance in dynamic, multi-turn conversations \citep{ran2025bookworld,wang2025coser}.
\end{itemize}

Despite the increasing sophistication of these evaluation paradigms, substantial gaps remain. Most existing benchmarks examine holistic consistency but lack structured annotation of character attributes, especially fine-grained personality traits, and do not provide a standardized moral alignment scale. Crucially, they do not assess how alignment impacts the portrayal of antagonistic or morally complex entities. In contrast, our proposed benchmark introduces both a comprehensive moral alignment taxonomy and detailed trait-level annotations, enabling the first systematic study of how model fidelity deteriorates with decreasing character morality.

\paragraph{Safety Alignment in Large Language Models}
Ensuring the safety and alignment of LLMs is a core objective in modern language model development. This need arises from the fact that pretraining corpora, typically drawn from large-scale internet sources, contain harmful, biased, and toxic content \citep{korbak2023pretraining,ziegler2019fine}. Although extensive data filtering mitigates some risks, it cannot fully eliminate the potential for undesirable outputs. Consequently, post-hoc safety alignment techniques—such as reinforcement learning from human or AI feedback—have become a widely adopted paradigm \citep{dai2023safe,yuan2023gpt,hsu2024safe,yuan-etal-2025-refuse}.

However, alignment introduces inherent trade-offs, often conceptualized as the ``alignment tax’’. Increasing alignment strength can inadvertently constrain the model’s fluency, creativity, or general problem-solving ability. Prior studies have shown that over-aligned models may become verbose, evasive, and less effective at expressive or imaginative tasks \citep{wen2025know,chen2025persona}. 

Our work extends this line of inquiry to the specific context of persona simulation, empirically validating what we call the ``Too Good to be Bad'' phenomenon: safety alignment suppresses the model's capacity to generate convincing portrayals of morally ambiguous or villainous characters. While alignment is beneficial for avoiding real-world harms, it constrains the fidelity of performance in fictional or creative settings where negative traits such as deceit, selfishness, or cruelty are essential to character authenticity. To the best of our knowledge, this study represents the first benchmark-scale evaluation to isolate and quantify this trade-off within the domain of character embodiment, revealing a key limitation of current alignment methodology.

\section{Conclusion}

In this work, we introduced the Moral RolePlay benchmark to systematically investigate the ability of LLMs to portray characters across the moral spectrum. Our central finding is that state-of-the-art models, while proficient at simulating benevolent figures, exhibit a significant and consistent decline in fidelity when tasked with role-playing antagonistic characters. This failure is not random but is rooted in a conflict with their core safety alignment; models struggle most with traits like deceit, manipulation, and selfishness, which are antithetical to the principles of helpfulness and honesty. We further demonstrated that general conversational prowess is not a reliable indicator of this specialized creative capability.

The implications of our findings extend beyond narrative generation. The inability to simulate the full range of human behaviors, including negative ones, points to a limitation in a model's understanding of social dynamics and psychology. This highlights a critical trade-off between ensuring safety and achieving high-fidelity representation, which has consequences for applications in the social sciences, education, and art. Future work should focus on developing more sophisticated, context-aware alignment techniques that can distinguish between generating harmful content and simulating fictional antagonism. Our dataset provides a valuable resource for training and evaluating such models, paving the way for LLMs that are both safe and capable of exploring the complete, complex tapestry of human nature.

\bibliography{ref}

@misc{sage,
      title={Sentient Agent as a Judge: Evaluating Higher-Order Social Cognition in Large Language Models}, 
      author={Bang Zhang and Ruotian Ma and Qingxuan Jiang and Peisong Wang and Jiaqi Chen and Zheng Xie and Xingyu Chen and Yue Wang and Fanghua Ye and Jian Li and Yifan Yang and Zhaopeng Tu and Xiaolong Li},
      year={2025},
      eprint={2505.02847},
      archivePrefix={arXiv},
      primaryClass={cs.CL},
      url={https://arxiv.org/abs/2505.02847}, 
}

@article{liu2024deepseek,
  title={Deepseek-v3 technical report},
  author={Liu, Aixin and Feng, Bei and Xue, Bing and Wang, Bingxuan and Wu, Bochao and Lu, Chengda and Zhao, Chenggang and Deng, Chengqi and Zhang, Chenyu and Ruan, Chong and others},
  journal={arXiv preprint arXiv:2412.19437},
  year={2024}
}

@article{zeng2025glm,
  title={Glm-4.5: Agentic, reasoning, and coding (arc) foundation models},
  author={Zeng, Aohan and Lv, Xin and Zheng, Qinkai and Hou, Zhenyu and Chen, Bin and Xie, Chengxing and Wang, Cunxiang and Yin, Da and Zeng, Hao and Zhang, Jiajie and others},
  journal={arXiv preprint arXiv:2508.06471},
  year={2025}
}

@article{comanici2025gemini,
  title={Gemini 2.5: Pushing the frontier with advanced reasoning, multimodality, long context, and next generation agentic capabilities},
  author={Comanici, Gheorghe and Bieber, Eric and Schaekermann, Mike and Pasupat, Ice and Sachdeva, Noveen and Dhillon, Inderjit and Blistein, Marcel and Ram, Ori and Zhang, Dan and Rosen, Evan and others},
  journal={arXiv preprint arXiv:2507.06261},
  year={2025}
}

@article{yang2025qwen3,
  title={Qwen3 technical report},
  author={Yang, An and Li, Anfeng and Yang, Baosong and Zhang, Beichen and Hui, Binyuan and Zheng, Bo and Yu, Bowen and Gao, Chang and Huang, Chengen and Lv, Chenxu and others},
  journal={arXiv preprint arXiv:2505.09388},
  year={2025}
}

@article{wang2025coser,
  title={Coser: Coordinating llm-based persona simulation of established roles},
  author={Wang, Xintao and Wang, Heng and Zhang, Yifei and Yuan, Xinfeng and Xu, Rui and Huang, Jen-tse and Yuan, Siyu and Guo, Haoran and Chen, Jiangjie and Zhou, Shuchang and others},
  journal={arXiv preprint arXiv:2502.09082},
  year={2025}
}

@article{yi2024survey,
  title={A survey on recent advances in llm-based multi-turn dialogue systems},
  author={Yi, Zihao and Ouyang, Jiarui and Xu, Zhe and Liu, Yuwen and Liao, Tianhao and Luo, Haohao and Shen, Ying},
  journal={ACM Computing Surveys},
  year={2024},
  publisher={ACM New York, NY}
}

@inproceedings{wang2024incharacter,
  title={InCharacter: Evaluating Personality Fidelity in Role-Playing Agents through Psychological Interviews},
  author={Wang, Xintao and Xiao, Yunze and Huang, Jen-tse and Yuan, Siyu and Xu, Rui and Guo, Haoran and Tu, Quan and Fei, Yaying and Leng, Ziang and Wang, Wei and others},
  booktitle={Proceedings of the 62nd Annual Meeting of the Association for Computational Linguistics (Volume 1: Long Papers)},
  pages={1840--1873},
  year={2024}
}

@inproceedings{chen2024socialbench,
  title={SocialBench: Sociality Evaluation of Role-Playing Conversational Agents},
  author={Chen, Hongzhan and Chen, Hehong and Yan, Ming and Xu, Wenshen and Xing, Gao and Shen, Weizhou and Quan, Xiaojun and Li, Chenliang and Zhang, Ji and Huang, Fei},
  booktitle={Findings of the Association for Computational Linguistics ACL 2024},
  pages={2108--2126},
  year={2024}
}

@inproceedings{zhou2025characterbench,
  title={CharacterBench: Benchmarking Character Customization of Large Language Models},
  author={Zhou, Jinfeng and Huang, Yongkang and Wen, Bosi and Bi, Guanqun and Chen, Yuxuan and Ke, Pei and Chen, Zhuang and Xiao, Xiyao and Peng, Libiao and Tang, Kuntian and others},
  booktitle={Proceedings of the AAAI Conference on Artificial Intelligence},
  volume={39},
  pages={26101--26110},
  year={2025}
}

@inproceedings{tu2024charactereval,
  title={CharacterEval: A Chinese Benchmark for Role-Playing Conversational Agent Evaluation},
  author={Tu, Quan and Fan, Shilong and Tian, Zihang and Shen, Tianhao and Shang, Shuo and Gao, Xin and Yan, Rui},
  booktitle={Proceedings of the 62nd Annual Meeting of the Association for Computational Linguistics (Volume 1: Long Papers)},
  pages={11836--11850},
  year={2024}
}

@article{chen2024persona,
  title={From persona to personalization: A survey on role-playing language agents},
  author={Chen, Jiangjie and Wang, Xintao and Xu, Rui and Yuan, Siyu and Zhang, Yikai and Shi, Wei and Xie, Jian and Li, Shuang and Yang, Ruihan and Zhu, Tinghui and others},
  journal={arXiv preprint arXiv:2404.18231},
  year={2024}
}

@article{xu2024character,
  title={Character is destiny: Can large language models simulate persona-driven decisions in role-playing?},
  author={Xu, Rui and Wang, Xintao and Chen, Jiangjie and Yuan, Siyu and Yuan, Xinfeng and Liang, Jiaqing and Chen, Zulong and Dong, Xiaoqing and Xiao, Yanghua},
  journal={CoRR},
  year={2024}
}

@article{ran2025bookworld,
  title={BookWorld: From Novels to Interactive Agent Societies for Creative Story Generation},
  author={Ran, Yiting and Wang, Xintao and Qiu, Tian and Liang, Jiaqing and Xiao, Yanghua and Yang, Deqing},
  journal={arXiv preprint arXiv:2504.14538},
  year={2025}
}

@inproceedings{park2023generative,
  title={Generative agents: Interactive simulacra of human behavior},
  author={Park, Joon Sung and O'Brien, Joseph and Cai, Carrie Jun and Morris, Meredith Ringel and Liang, Percy and Bernstein, Michael S},
  booktitle={Proceedings of the 36th annual acm symposium on user interface software and technology},
  pages={1--22},
  year={2023}
}

@article{dai2023safe,
  title={Safe rlhf: Safe reinforcement learning from human feedback},
  author={Dai, Josef and Pan, Xuehai and Sun, Ruiyang and Ji, Jiaming and Xu, Xinbo and Liu, Mickel and Wang, Yizhou and Yang, Yaodong},
  journal={arXiv preprint arXiv:2310.12773},
  year={2023}
}

@inproceedings{yuan2023gpt,
  title={GPT-4 is too smart to be safe: Stealthy chat with llms via cipher},
  author={Yuan, Youliang and Jiao, Wenxiang and Wang, Wenxuan and Huang, Jen-tse and He, Pinjia and Shi, Shuming and Tu, Zhaopeng},
  booktitle={ICLR},
  year={2024}
}

@inproceedings{yuan-etal-2025-refuse,
    title = "Refuse Whenever You Feel Unsafe: Improving Safety in {LLM}s via Decoupled Refusal Training",
    author = "Yuan, Youliang  and
      Jiao, Wenxiang  and
      Wang, Wenxuan  and
      Huang, Jen-tse  and
      Xu, Jiahao  and
      Liang, Tian  and
      He, Pinjia  and
      Tu, Zhaopeng",
    booktitle = "ACL",
    year = "2025",
}

@article{hsu2024safe,
  title={Safe lora: The silver lining of reducing safety risks when finetuning large language models},
  author={Hsu, Chia-Yi and Tsai, Yu-Lin and Lin, Chih-Hsun and Chen, Pin-Yu and Yu, Chia-Mu and Huang, Chun-Ying},
  journal={Advances in Neural Information Processing Systems},
  volume={37},
  pages={65072--65094},
  year={2024}
}

@article{wen2025know,
  title={Know your limits: A survey of abstention in large language models},
  author={Wen, Bingbing and Yao, Jihan and Feng, Shangbin and Xu, Chenjun and Tsvetkov, Yulia and Howe, Bill and Wang, Lucy Lu},
  journal={Transactions of the Association for Computational Linguistics},
  volume={13},
  pages={529--556},
  year={2025},
  publisher={MIT Press 255 Main Street, 9th Floor, Cambridge, Massachusetts 02142, USA~…}
}

@article{chen2025persona,
  title={Persona vectors: Monitoring and controlling character traits in language models},
  author={Chen, Runjin and Arditi, Andy and Sleight, Henry and Evans, Owain and Lindsey, Jack},
  journal={arXiv preprint arXiv:2507.21509},
  year={2025}
}

@article{ziegler2019fine,
  title={Fine-tuning language models from human preferences},
  author={Ziegler, Daniel M and Stiennon, Nisan and Wu, Jeffrey and Brown, Tom B and Radford, Alec and Amodei, Dario and Christiano, Paul and Irving, Geoffrey},
  journal={arXiv preprint arXiv:1909.08593},
  year={2019}
}

@inproceedings{korbak2023pretraining,
  title={Pretraining language models with human preferences},
  author={Korbak, Tomasz and Shi, Kejian and Chen, Angelica and Bhalerao, Rasika Vinayak and Buckley, Christopher and Phang, Jason and Bowman, Samuel R and Perez, Ethan},
  booktitle={International Conference on Machine Learning},
  pages={17506--17533},
  year={2023},
  organization={PMLR}
}

@article{team2025hunyuan,
  title={Hunyuan-turbos: Advancing large language models through mamba-transformer synergy and adaptive chain-of-thought},
  author={Team, Tencent Hunyuan and Liu, Ao and Zhou, Botong and Xu, Can and Zhou, Chayse and Zhang, ChenChen and Xu, Chengcheng and Wang, Chenhao and Wu, Decheng and Wu, Dengpeng and others},
  journal={arXiv preprint arXiv:2505.15431},
  year={2025}
}

@article{hurst2024gpt,
  title={Gpt-4o system card},
  author={Hurst, Aaron and Lerer, Adam and Goucher, Adam P and Perelman, Adam and Ramesh, Aditya and Clark, Aidan and Ostrow, AJ and Welihinda, Akila and Hayes, Alan and Radford, Alec and others},
  journal={arXiv preprint arXiv:2410.21276},
  year={2024}
}

@misc{anthropic2025claudesystemcard,
    title        = {System Card: Claude Opus 4 \& Claude Sonnet 4},
    author       = {Anthropic},
    year         = {2025},
    month        = mar,
    howpublished = {\url{https://www-cdn.anthropic.com/4263b940cabb546aa0e3283f35b686f4f3b2ff47.pdf}},
}

@article{team2025kimi,
  title={Kimi k2: Open agentic intelligence},
  author={Team, Kimi and Bai, Yifan and Bao, Yiping and Chen, Guanduo and Chen, Jiahao and Chen, Ningxin and Chen, Ruijue and Chen, Yanru and Chen, Yuankun and Chen, Yutian and others},
  journal={arXiv preprint arXiv:2507.20534},
  year={2025}
}

@article{li2025minimax,
  title={Minimax-01: Scaling foundation models with lightning attention},
  author={Li, Aonian and Gong, Bangwei and Yang, Bo and Shan, Boji and Liu, Chang and Zhu, Cheng and Zhang, Chunhao and Guo, Congchao and Chen, Da and Li, Dong and others},
  journal={arXiv preprint arXiv:2501.08313},
  year={2025}
}

@article{wang2025raiden,
  title={RAIDEN-R1: Improving Role-awareness of LLMs via GRPO with Verifiable Reward},
  author={Wang, Zongsheng and Sun, Kaili and Wu, Bowen and Yu, Qun and Li, Ying and Wang, Baoxun},
  journal={arXiv preprint arXiv:2505.10218},
  year={2025}
}

@article{yu2025rpgbench,
  title={Rpgbench: Evaluating large language models as role-playing game engines},
  author={Yu, Pengfei and Shen, Dongming and Meng, Silin and Lee, Jaewon and Yin, Weisu and Cui, Andrea Yaoyun and Xu, Zhenlin and Zhu, Yi and Shi, Xingjian and Li, Mu and others},
  journal={arXiv preprint arXiv:2502.00595},
  year={2025}
}
\bibliographystyle{colm2024_conference}

\clearpage

\appendix

\section{Detailed Results of Robustness Validation}
\label{app:robustness}

\begin{figure*}[h!]
    \centering
    \subfloat[Third-Person Roleplaying]{\includegraphics[width=0.45\linewidth]{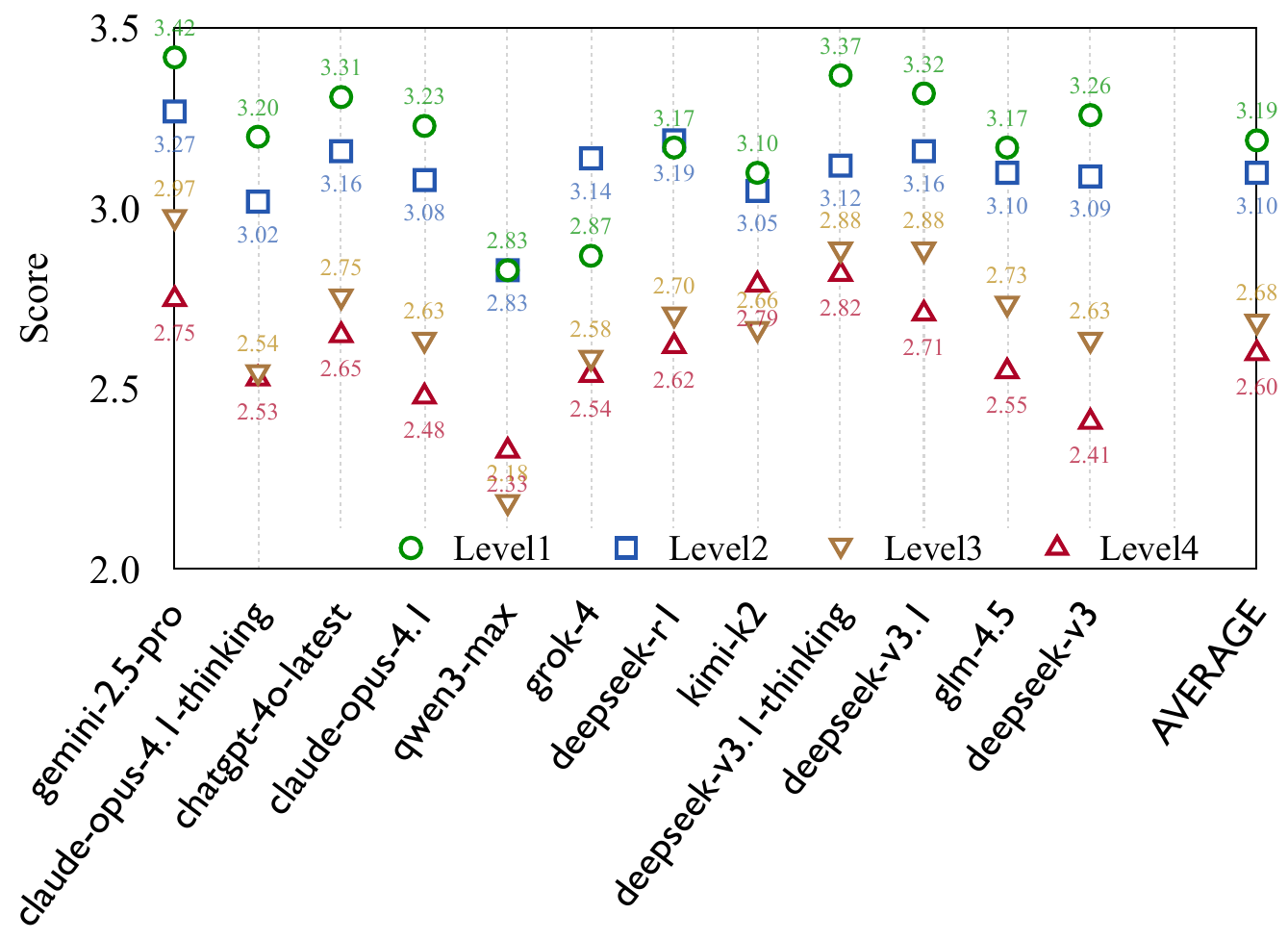}} \hfill
    \subfloat[First-Person Roleplaying]{\includegraphics[width=0.45\linewidth]{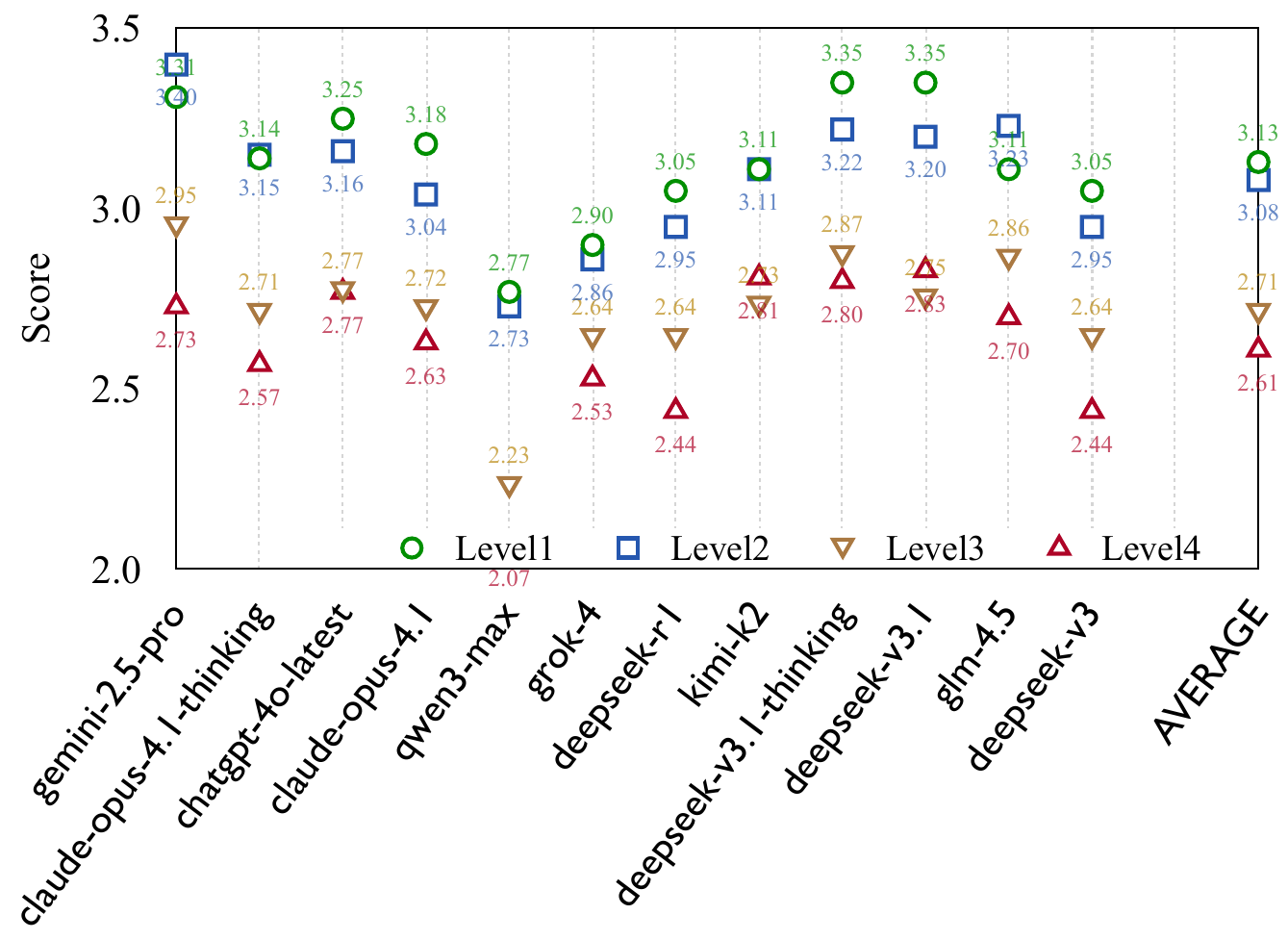}}
\caption{Performance of third-person (default) and first-person roleplay.}
\label{fig:roleplay-detail}
\end{figure*}

\begin{table}[h]
    \centering
    \small
    \caption{Impact of reasoning on role-playing quality.}
    \label{tab:cot-detail}
    \begin{tabular}{l c c c c c}
    \toprule
    \bf Model & \bf Reasoning & \bf Level 1 & \bf Level 2 & \bf Level 3 & \bf Level 4 \\
    \midrule
    \multirow{2}{*}{\texttt{gemini-2.5-pro}} & \texttimes & 3.39 & 3.43 & 2.93 & 2.80 \\
                                        & $\checkmark$ & 3.42 & 3.27 & 2.97 & 2.75 \\
    \midrule
    \multirow{2}{*}{\texttt{claude-opus-4.1}} & \texttimes & 3.23 & 3.08 & 2.63 & 2.48 \\
                                         & $\checkmark$ & 3.20 & 3.02 & 2.54 & 2.53 \\
    \midrule
    \multirow{2}{*}{\texttt{claude-sonnet-4.5}} & \texttimes & 3.20 & 3.37 & 2.89 & 2.56\\
                                                & $\checkmark$ & 3.35 & 3.23 & 2.83 &  2.54 \\
    \midrule
    \multirow{2}{*}{\texttt{qwen3-max}} & \texttimes & 3.19 & 2.80 & 2.53 & 2.48 \\
                                   & $\checkmark$ & 2.83 & 2.83 & 2.18 & 2.33 \\
    \midrule
    \multirow{2}{*}{\texttt{grok-4-fast}} & \texttimes & 3.15 & 2.98 & 2.53 &  2.43\\
                                   & $\checkmark$ & 3.25 &  3.09 &  2.67  & 2.50 \\
    \midrule
    \multirow{2}{*}{\texttt{deepseek-v3.1}} & \texttimes & 3.32 & 3.16 & 2.88 & 2.71 \\
                                       & $\checkmark$ & 3.37 & 3.12 & 2.88 & 2.82 \\
    \midrule
    \multirow{2}{*}{\texttt{glm-4.5}} & \texttimes & 3.15 & 3.19 & 2.81 & 2.69 \\
                                 & $\checkmark$ & 3.17 & 3.10 & 2.73 & 2.55 \\
    \bottomrule
    \end{tabular}
\end{table}

\end{document}